# Seeding the Singularity for A.I.

Pavel Kraikivski
Academy of Integrated Science, Division of Systems Biology, Virginia Tech

"The singularity" refers to an idea that once a machine having an artificial intelligence surpassing the human intelligence capacity is created, it will trigger explosive technological and intelligence growth. I propose to test the hypothesis that machine intelligence capacity can grow autonomously starting with an intelligence comparable to that of bacteria - microbial intelligence. The goal will be to demonstrate that rapid growth in intelligence capacity can be realized at all in artificial computing systems. I propose the following three properties that may allow an artificial intelligence to exhibit a steady growth in its intelligence capacity: (i) learning with the ability to modify itself when exposed to more data, (ii) acquiring new functionalities (skills), and (iii) expanding or replicating itself. The algorithms must demonstrate a rapid growth in skills of data-processing and analysis and gain qualitatively different functionalities, at least until the current computing technology supports their scalable development. The existing algorithms that already encompass some of these or similar properties, as well as missing abilities that must yet be implemented, will be reviewed in this work. Future computational tests could support or oppose the hypothesis that artificial intelligence can potentially grow to the level of superintelligence which overcomes the limitations in hardware by producing necessary processing resources or by changing the physical realization of computation from using chip circuits to using quantum computing principles.

**Introduction**

"The rapid growth" is the classification term that is notably and widely used in economics [1, 2], population dynamics and industrial revolution studies [3], computer sciences [4, 5] and science's genealogy [6] when the abnormally high growth of something in the corresponding field is observed. It has been hypothesized that an intense growth of intelligence will occur when an artificial intelligence surpasses the human intelligence power. This phenomenon is termed "the intelligence explosion" or "the singularity" [7-10]. At a singularity point, the slope of the intelligence growth curve must diverge, approaching an infinite value, see Figure 1A. However, it is not obvious that the singularity and explosive intelligence growth will occur at any future time. It might even be possible that the intelligence evolution will exhibit a series of growth rate jumps interrupted by periods of stagnation, thereby having multiple singularities at different times. The explosive intelligence growth hypothesis can be tested by exploring the autonomous artificial intelligence algorithms that are able to grow their intellectual capabilities without human assistance. Autonomous algorithms could exhibit different growth regimes in intelligence capacity, for some examples see Figure 1B, thus exploration of dynamic behaviors of autonomous algorithms could help us to understand possible scenarios of artificial intelligence evolution. The different growth regimes could be investigated using computational modeling, simulating different strategies of intelligence capacity growth. In this manuscript, I will propose and discuss properties that algorithms must have in order to realize the computational investigation of autonomous artificial intelligence evolution.

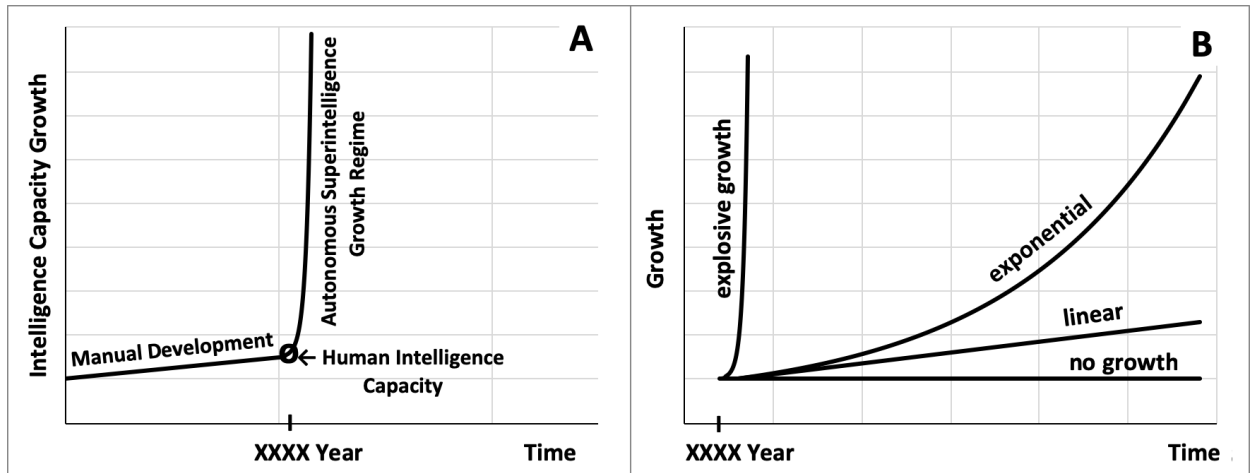

Figure 1. Hypothetical scenarios of artificial intelligence capacity growth. (A) demonstrates the hypothesis of singularity occurring at a time when artificial intelligence reaches the human intelligence capacity. The artificial intelligence exhibits explosive growth developing and evolving independently of humans. (B) shows other possible growth regimes that can be explored by simulating artificial intelligence growth with algorithms capable of learning, gaining new skills, and replicating.

The intelligence of a living organism or a machine can be described as its ability to digest perceived information into different forms of knowledge and skills which are then applied towards adaptive behaviors in changing circumstances. Progress in machine learning and advances in technology of developing and building multicore processors and graphical processing units have made it practicable to apply AI computing techniques to such remarkable projects as precision medicine [11], stock trading software robots [12] and driverless cars [13]. However, despite the fact that machines can classify enormous amounts of data and be accurate in making predictions, they are not well-designed and skillful enough to do jobs that even simple microbial cells can do. Bacteria and other simple single-cell organisms, for example, demonstrate remarkable abilities to organize themselves in alternating circumstances by working out things for themselves [14]. Therefore, to catch up with living organisms, machines have to pursue a goal to gain new skills and functionalities that will allow their processing units to solve qualitatively different tasks.

Perhaps driven by economical and security reasons, we have been designing machines with architecture, interfaces and software that support human-machine partnerships rather than supporting an architecture where a program itself could replicate, spread and gain new functionalities without our full control and assistance. The goal of this manuscript is to advocate that granting this freedom to an algorithm is necessary for an artificial intelligence to demonstrate a steady growth in intelligence capacity and thus potentially to reach the power of superintelligence.

**Learning Ability**

An important part of an artificial intelligence paradigm is machine-learning, referring to the study and design of computer algorithms that have potential to be trained to identify patterns and correlations in large input data-sets. Machine-learning algorithms incorporate the ability of the program to modify its output when exposed to more data [15]. The classification of the data into a set of patterns allows a machine to make predictions or decisions when the machine receives a similar kind of data. Machine-learning has received great attention and investments from such tech giants as Google Inc. (developing machine learning system called TensorFlow [16, 17]), Facebook Inc. [18], Amazon Inc. (funding a conversational AI project called Alexa [19]), Apple Inc. [20], Microsoft Inc. (investing in Element AI [21]), and other companies that are able to collect Big Data requiring a tool to efficiently perform classification of these data. Overall, the machine's ability to learn from data is already well-implemented and widely used for many applications.

Machine-learning algorithms use two distinct techniques called supervised and unsupervised learning [22-24]. Supervised learning is typically used to map input to a set of training examples, thus solving a classification task [22, 23]. For example, a computing system might be provided with labeled images of fruits in categories such as apples, pears, plums, etc. The number of training examples should be sufficient to complete the learning so that the system is able to successfully categorize unlabeled pictures of fruits. Unsupervised learning is used to uncover the inherent structure of data without trying to explicitly label possible outcomes [20, 21, 24]. It can be used for representation learning or splitting data into distinct clusters or patterns and to discover rules that describe large portions of data. For example, unsupervised learning algorithms can cluster consumers into groups based on their similar product preferences. Consumers who buy iPhones, for instance, also tend to buy Mac computers and prefer to pay with Amex credit cards. In representation learning, the unsupervised methods can find hidden relationships in data which reduce the dimensionality of the data. This method could also find some features describing the data to be redundant. For example, whether consumers are tall or short could be redundant information for predicting a payment method used by consumers.

Supervised and unsupervised machine-learning algorithms are often realized using artificial neural network frameworks [25, 26]. The artificial neural networks have nodes (artificial neurons) and connections between nodes (artificial synapses) which imitate biological neural networks. An adjustable parameter value is assigned to each connection between nodes. The numerical value of the parameter represents the strength of the connection and can be changed during the network training process. Often, nodes are organized in layers and each node in a layer is connected to some nodes in other layers. However, many different architectures of artificial neural networks are used depending on a task [27-31]. An essential learning feature of artificial neural networks is that networks modify themselves. Their internal parameters, when exposed to more data [32, 33], mimic the learning mechanisms of biological brains wherein synaptic connections among neurons are strengthen when signals propagate through connected neurons.

Machine-learning techniques are proven to be effective in solving scientific and medical problems [11], driving cars [13], financial trading [12] and many other applications. However, despite the fact that machine-learning algorithms can learn how to classify data by applying a specific method, they have an obvious disadvantage compared to living organisms.  Current machine-learning programs do not have the goal of learning new methods of building or incorporating tools and techniques which can be applied to solve qualitatively different kinds of problems.

**Acquiring New Skills**

An artificial intelligence should be able to acquire new skills. As living organisms that might evolve into a new species when they acquire new skills enabling them to better survive, reproduce and become more intelligent than successive generations, artificial intelligence must have such potential, as well. In nature, the ability of an organism to develop qualitatively new skills often comes from random, favorable genetic mutations that rarely occur or due to an exchange of genetic material between different organisms [34]. However, algorithms could "evolve" by utilizing already existing functions and methods as building blocks, endeavoring to solve problems. Algorithms could exchange methods and functions among each other, and combine them to build new skills. In addition, programs must have goals to develop, refine and improve tools to analyze data.

Once this is accomplished, computer programs can be provided with building blocks that can be combined into comprehensive algorithms expanding their potential for machine-learning and other intelligence skills. A good example of this is expanding an algorithm with a simple one-layer-neural-network architecture to a multilayer-deep-learning-network. Here, the building blocks are layers containing artificial neurons. Adding such processing layers allows the machine to learn complex features and solve complex problems, thereby greatly improving the machine's learning ability [35].  Complex biological and medical data can be explained by combining three different machine-learning methods: Artificial Neural Network, Random Forest, and Support Vector Machines [36]. Building blocks can therefore be elementary functions, single artificial neurons, layers of neurons, or comprehensive machine-learning methods. Ultimately, the combination of qualitatively different techniques, functions and methods may provide a machine with more powerful tools. Moreover, a machine-learning algorithm that can first learn to write "if-then" statements and implement rules of logic and symbolic reasoning will be then be able to code and develop itself.

There are various machine-learning methods [37-41], and artificial intelligence algorithms that are realized in such impressive applications as conversational AI [19], text-to-speech neural converter [27], reinforcement learning systems that learn from their own experience [42], automatic text mining algorithms [43], and algorithms that can code [44]. However, these comprehensive algorithms do not set a goal for themselves to acquire new skills, therefore they would not be able to demonstrate autonomous growth of their intelligence capacity. Autonomous algorithms that work beyond simple neural network frameworks, have goals to evolve and expand their functionalities and adopt and test new methods without human assistance could be used to explore the autonomous evolution of artificial intelligence capacity. The challenging part will be to provide these algorithms with freedom of expansion and replication, thus not limiting the evolution rate of the autonomous algorithms.

**Replication**

Undoubtedly, the intelligence capacity growth of machines can be enhanced by creating a large population of algorithms that independently evolve and acquire new skills. This can be achieved by allowing algorithms to replicate and spread copies of themselves, which might also be modified copies. Replication techniques for computer programs have a long usage history and were implemented since the beginning of the computer era. Formally, all software users are aware of existence of malicious software (malware) that can infect computers and propagate themselves. Malware that have the ability to self-replicate and spread copies of themselves are called computer viruses and worms. A computer virus propagates by copying itself into another program and depends on the host program to spread itself, whereas a worm program can operate independently of other files. While malicious software is certainly undesirable, it is a good demonstration of the technology of algorithm replication if harmful parts of the malware are removed.

The beginning of life is marked when a simple molecule has gained the ability to reproduce itself [45]. Reproduction is certainly essential for the evolution of living species. In nature, living things use either sexual or asexual strategies for producing offspring. In asexual reproduction, offspring are genetically identical to a parent. In sexual reproduction, offspring inherit parts of genetic information from both parents. Computer viruses and worms mimic asexual reproduction. However, algorithms may be designed to resemble a sexual reproduction when two algorithms share, merge and mix their functions to generate more variable and functional offspring-algorithms. The ability to share and mix functions and to subsequently replicate can positively speed up the evolution of algorithms.

Algorithms can utilize other evolution techniques that have been proven to be useful for living things. For example, a whole-genome duplication played an important role in the evolution of living organisms [46]. Genome duplication provides raw genetic material and, with it, leads to opportunities for some advantageous duplicated genes to survive and acquire novel functions. By mimicking this time-tested propensity of living things, algorithms could also perform duplication of functions that can be then used by the algorithm to create novel or modified functions. The internal duplication of functions by a program is a way for an algorithm to improve existing methods or to acquire new computing techniques without installing new or additional versions of the program. Overall, nature and living organisms demonstrate various evolution, survival and reproduction strategies that teach us how to design rules that could accelerate the evolution of algorithms. These time-tested techniques used by living things could be implemented into the autonomous algorithms that will be able to produce more variable offspring-algorithms, thus enhancing artificial intelligence evolution.

**Discussion**

In this manuscript, it is argued that it is important to apply current computational technologies and modeling to test the hypothesis that machine intelligence capacity can autonomously grow by acquiring new skills and propagating itself. A human role should be limited by seeding the initial rules for the intelligence growth and providing initial computational resources. It has been proposed that fast intelligence growth will be possible only when an artificial intelligence will reach a human intelligence capacity [7-9]. However, my assumption is that machine intelligence can start growing with an initial intelligence capacity that is substantially lower than human intelligence if certain rules for the intelligence capacity growth are implemented in the machine algorithms. These rules include the ability of an algorithm to learn patterns from data, acquire new learning skills and replicate itself. The current computing technologies support the implementation of all of these rules in the form of autonomous algorithms.

The ability of algorithms to increase their intelligence capacity can be simulated and explored using computational modeling techniques. Computational modeling is widely used in science to understand dynamic behaviors of complex systems. This modeling may often unravel counterintuitive dynamic behaviors of the systems. Also, the modeling results can be understood by exploring the system's behavior in all possible regimes and influences of different parameters. Therefore, computational modeling could be a suitable tool to explore different strategies and regimes for growth of artificial intelligence capacity.

As proposed by philosophers, the superintelligence will have a great impact on society, bringing potential positive benefits as well as possible dangers [9, 10]. Thus, to predict and be able to influence the possible outcomes of explosive intelligence growth, it is now important to test the autonomous artificial intelligence growth using current advanced computing methods. Modeling predictions of possible outcomes will help us to understand the impact of the fast growth of artificial intelligence capacity on our society and find ways to maximize the chances of a good outcome from the artificial intelligence evolution.

Current machine-learning algorithms are already strong in retaining information and recognizing patterns in Big Data but weak in deducing new skills and knowledge from it. If artificial intelligence will only gain a high capacity in logic, planning and problem-solving, it will still lag behind the human model without the capability of understanding, critical thinking and self-awareness, emotional knowledge, creativity, and consciousness that humans possess. Therefore, building algorithms that are capable of conscious interpretation of perceived information [47, 48] will be crucial for an artificial intelligence to be able to surpass the human intelligence capacity. However, as the first step, we could start exploring the possible outcomes of autonomous intelligence growth by allowing existing algorithms to gain new skills and replicate themselves.